\documentclass[conference]{IEEEtran}
\IEEEoverridecommandlockouts
\usepackage{cite}
\usepackage{amsmath,amssymb,amsfonts}
\usepackage{algorithmic}
\usepackage{graphicx}
\usepackage{textcomp}
\usepackage{xcolor}
\usepackage{hyperref}
\usepackage{tabularx}
\usepackage{booktabs}
\usepackage{multirow}

\def\BibTeX{{\rm B\kern-.05em{\sc i\kern-.025em b}\kern-.08em
    T\kern-.1667em\lower.7ex\hbox{E}\kern-.125emX}}
\begin{document}

\title{ResNetVLLM - Multi-modal Vision LLM for the Video Understanding Task \\
\thanks{University of Windsor.}
}

% \author{\IEEEauthorblockN{Anonymous Authors}}
\author{\IEEEauthorblockN{Ahmad Khalil}
\IEEEauthorblockA{\textit{Dept of Computer Science} \\
\textit{University of Windsor}\\
Windsor, Canada \\
https://orcid.org/0009-0009-4483-5839}
\and
\IEEEauthorblockN{Mahmoud Khalil}
\IEEEauthorblockA{\textit{Dept of Computer Science} \\
\textit{University of Windsor}\\
Windsor, Canada \\
https://orcid.org/0009-0001-4112-0848}
\and
\IEEEauthorblockN{Alioune Ngom}
\IEEEauthorblockA{\textit{Dept of Computer Science} \\
\textit{University of Windsor}\\
Windsor, Canada \\
angom@uwindsor.ca}
}

\maketitle

\begin{abstract}
In this paper, we introduce ResNetVLLM (ResNet Vision LLM), a novel cross-modal framework for zero-shot video understanding that integrates a ResNet-based visual encoder with a Large Language Model (LLM. ResNetVLLM addresses the challenges associated with zero-shot video models by avoiding reliance on pre-trained video understanding models and instead employing a non-pretrained ResNet to extract visual features. This design ensures the model learns visual and semantic representations within a unified architecture, enhancing its ability to generate accurate and contextually relevant textual descriptions from video inputs. Our experimental results demonstrate that ResNetVLLM achieves state-of-the-art performance in zero-shot video understanding (ZSVU) on several benchmarks, including MSRVTT-QA, MSVD-QA, TGIF-QA FrameQA, and ActivityNet-QA.
\end{abstract}


\section{Introduction}
Large language models (LLMs) \cite{touvron2023llama2, touvron2023llama, taori2023stanford, openai2023gpt4, openai2023chatgpt, chiang2023vicuna} have advanced natural language understanding tasks, demonstrating exceptional abilities in comprehending human intentions and interactions. Building on the progress of LLMs, multi-modal LLMs (MLLMs) \cite{dai2023instructblip, gao2023llamaadapter, liu2023visual, zhu2023minigpt4} have furthered vision-language learning by integrating visual encoders with LLMs and fine-tuning them on language-image instruction-following data.

Recently, there has been a surge in video understanding models that leverage LLMs. These models are capable of performing a variety of video understanding tasks. A notable advantage of using LLMs in video understanding is the ease of integrating existing LLM frameworks. This integration enables the reuse of LLM infrastructure and leverages the optimizations developed by the AI community over many years. While LLMs are widely used as foundational models across various modalities, visual encoders remain the predominant approach for video understanding. However, developing zero-shot video-based LLMs (VideoLLMs) remains a substantial challenge. These models must draw meaningful conclusions about unseen video content, relying solely on data from seen training concepts and supplementary high-level semantic label information \cite{sylvain2020locality}. Additionally, they must handle a substantial volume of tokens to model spatial-temporal dependencies across successive video frames simultaneously \cite{radford2021learning}.

\begin{figure}[t]
    \centering
    \includegraphics[height=0.20\textheight]{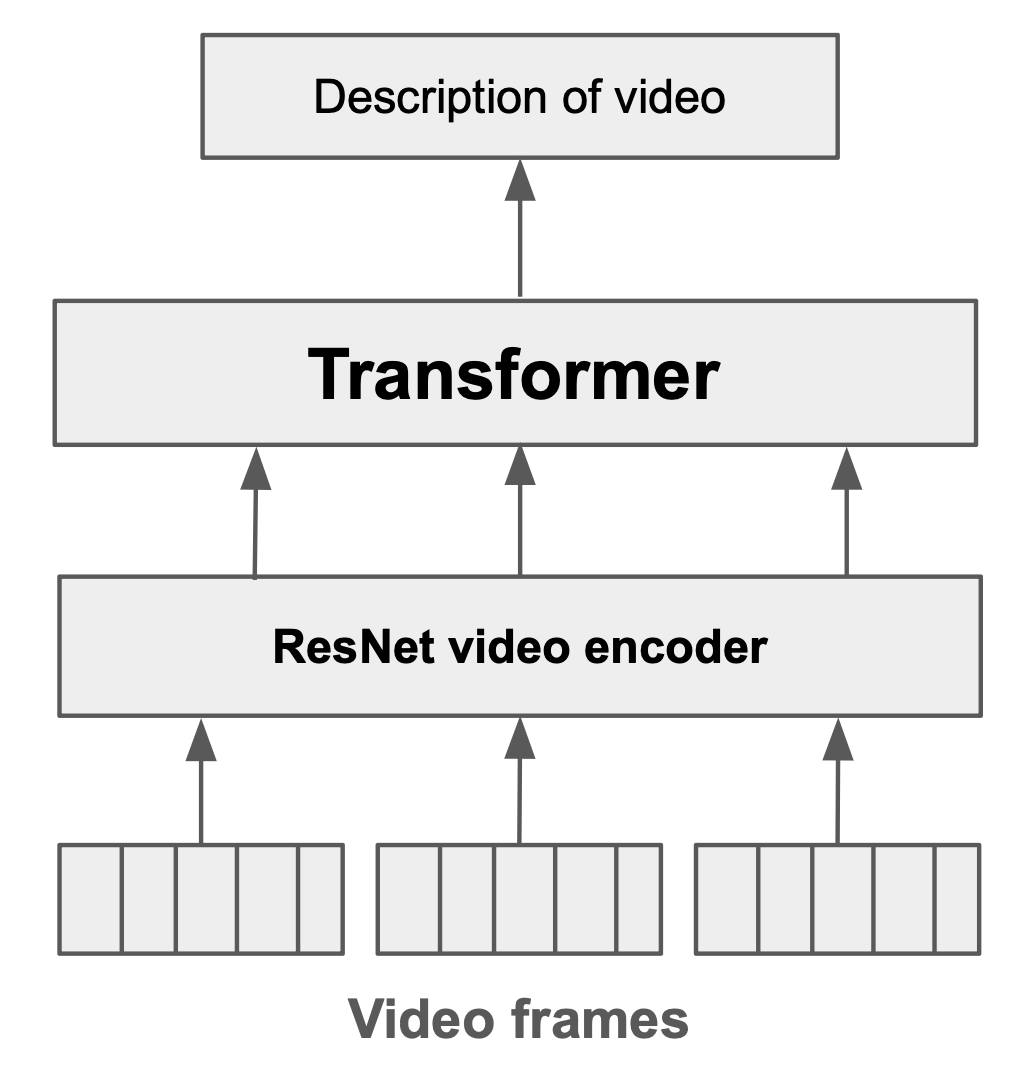}
    \caption{Overview of ResNetVLLM.}
    \label{fig:overview-ResNetVLLM}
\end{figure}

To tackle these challenges, recent methods in zero-shot VideoLLMs primarily leverage visual features obtained from pre-trained video understanding models. These methods use either pooling operations \cite{luo2023valley, maaz2023videochatgpt} or query aggregation \cite{zhang2023videollama, li2023videochat, song2023moviechat} on the token sequence of the video before passing them to the LLM. Nevertheless, these approaches have their limitations.

Visual features typically obtained from pre-trained video understanding models \cite{dutran2015learning, carreira2017quo} remain unchanged throughout the training process. As a result, they might lack sufficient information to acquire a comprehensive representation \cite{li2019joint}. Additionally, video content typically receive manual labeling, offering limited descriptions for many intricate actions, resulting in imprecise or incomplete semantic information \cite{pourpanah2020review, zhang2017learning}. For instance, numerous video content receive labels comprising minimal words (e.g., "play soccer," "jump in water," "skip," "run"). On the other hand, long-duration videos often encompass a multitude of pivotal actions, intricate activities, and camera motions (e.g., "The woman in the video is holding a paintbrush and palette, surrounded by a canvas and various tubes of paint. She wears a smock splattered with colorful pigments, indicating she's actively engaged in painting"). Consequently, breaking down intricate activities into sequences of smaller activities impacts the knowledge transferability from seen video content to unseen video content in video understanding \cite{pourpanah2020review, zhang2017learning}.

In this paper, we propose ResNetVLLM, a cross-modal framework based on ResNet Networks and VideoLLMs that is both straightforward and efficient for comprehensive video understanding. Rather than using pretrained feature extractors, ResNetVLLM incorporates a vanilla, non-pretrained ResNet module into a Transformer to extract visual features. This approach ensures that no prior knowledge of unseen video content is introduced during training. Additionally, ResNetVLLM combines the learning of visual representations and visual-semantic associations within a single unified architecture. This model design offers an inherent mechanism for learning visual and semantic representations within a shared knowledge space. This approach bridges the semantic gap, promoting the learned visual embeddings to be both discriminative and more semantically consistent.

The main contributions of this paper are:
\begin{itemize}
    \item We propose ResNetVLLM, a cross-modal framework based on ResNet Networks and LLMs that is both straightforward and efficient for comprehensive understanding of videos.
    \item Our framework achieves new state-of-the-art results for zero-shot video understanding (ZSVU) on the MSRVTT-QA \cite{xu2017video}, MSVD-QA \cite{xu2017video}, TGIF-QA FrameQA \cite{jang2017tgif}, and ActivityNet-QA \cite{yu2019activitynet} benchmarks.
\end{itemize}

\section{Related Work}
Any model that integrates written, spoken, signed, or other forms of human language into its training signal likely employs natural language as a supervisory source. This field is vast, encompassing a substantial portion of distributional semantics research. It includes topic models \cite{blei2003latent}, and vector representations for words, sentences, and paragraphs \cite{mikolov2013efficient,kiros2015skip,le2014distributed}, as well as language models \cite{bengio2003neural}.

\textbf{LLMs} have transformed natural language processing in recent years. Since language models can seamlessly incorporate multiple training tasks, selecting these tasks is crucial. Models like GPT-3 \cite{brown2020language} and PaLM \cite{chowdhery2022palm} demonstrate that training on a variety of tasks results in beneficial scaling effects for zero- and few-shot tasks. LLMs such as GPT-3 \cite{brown2020language}, OPT \cite{zhang2022opt}, and LLaMA \cite{touvron2023llama,touvron2023llama2} employ auto-regressive Transformer architectures to predict subsequent tokens, showcasing exceptional flexibility and generalization capabilities. These models demonstrate remarkable abilities in language generation and contextual learning, with the capacity to comprehend complex tasks based on user prompts in a zero-shot manner, highlighting their impressive flexibility and adaptability. The demonstrated capabilities of LLMs have motivated researchers to optimize their performance through fine-tuning.

\textbf{Vision Language Models} have driven notable advances in computer vision. These models represent a substantial step forward in developing versatile vision models capable of addressing multiple tasks concurrently \cite{radford2021learning,alayrac2022flamingo,gupta2022towards,maaz2022classagnostic}. One notable example is CLIP \cite{radford2021learning}, trained with 400 million image-text pairs, which has shown remarkable zero-shot performance across multiple benchmarks. It has been utilized in diverse applications beyond its original scope, including object detection and segmentation \cite{rasheed2022bridging,liang2023open}, and applications in 3D contexts \cite{rozenberszki2022language,ni2022expanding}. Several efforts have been undertaken to apply CLIP to video contexts \cite{liang2023open,wang2021actionclip}. For instance, ViFi-CLIP \cite{rasheed2023finetuned} proposes temporal pooling across video frames, similar to our approach, to adapt the CLIP model for video tasks.

\textbf{Video LLMs (VideoLLMs)} are generally developed from transformer-based language models, which effectively integrate multiple tasks during pre-training and exhibit strong zero-shot capabilities. Video-ChatGPT \cite{maaz2023videochatgpt} is a video LLM designed to align video representations with a LLM, thereby improving its capability to generate insightful discussions about video content. It leverages methodologies used in creating vision-language (VL) models tailored for the video domain. Due to the scarcity of video-caption pairs and the extensive resources needed to train from scratch, these models typically modify pretrained image-based VL models for video-related tasks \cite{ni2022expanding,wang2021actionclip,rasheed2023finetuned}. Video-ChatGPT utilizes a comparable strategy by using the Language-aligned Large Vision Assistant (LLaVA) \cite{liu2023visual} as its base model. LLaVA combines the visual encoder from CLIP \cite{radford2021learning} with the Vicuna language decoder \cite{chiang2023vicuna} and undergoes end-to-end fine-tuning on synthesized instructional vision-language datasets. This model is fine-tuned with video-instruction data, adapting it for video conversation tasks. The video-instruction data is sourced through a combination of manual and automated processes within the instruction generation framework. Adapting the model with video-specific instructions enables it to account for temporal dynamics, frame-to-frame consistency, and long-range relationships inherent in video data. Consequently, Video-ChatGPT \cite{maaz2023videochatgpt} excels in video reasoning, creativity, and comprehending spatial, temporal, and action-related elements within videos.

\section{Method}
\begin{figure*}[!t]
    \centering
    \includegraphics[width=350pt,height=120pt]{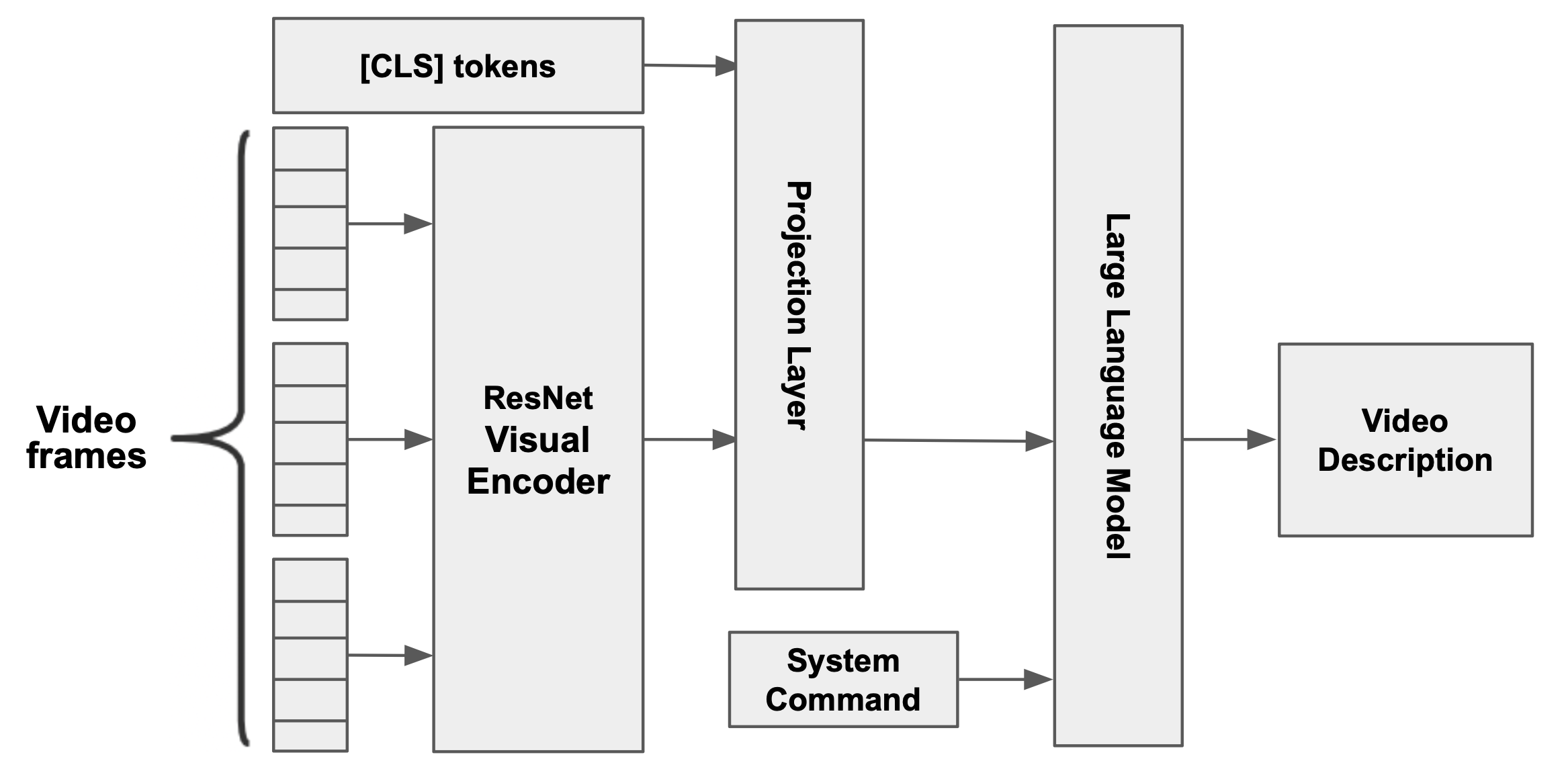}
    \caption{Overall architecture of the proposed ResNetVLLM framework.}
    \label{fig:details-ResNetVLLM}
\end{figure*}
ResNetVLLM is a sophisticated model designed to generate textual responses based on video inputs. It integrates video representations with natural language processing to facilitate tasks such as video understanding and descriptive caption generation.

ResNetVLLM combines a Residual Neural Network (ResNet) \cite{he2015deep} visual encoder with a GPT-based language model to generate text. The visual encoder processes video frames to capture spatiotemporal representations, which are then used by the language model to produce textual descriptions. ResNetVLLM is trained on extensive datasets of video-text pairs, with additional fine-tuning on specific tasks or datasets to enhance its performance. The training consists of two stages: warm-up and joint training. This sequential optimization approach targets different parts of the model at various stages. Figure \ref{fig:details-ResNetVLLM} illustrates the conceptual diagram of our framework.

This section elaborates on the approach used to develop ResNetVLLM. We first define the problem we address, then describe the ResNetVLLM model formulation, and finally discuss our strategy for zero-shot video understanding. Sub-section \ref{sec:problem_definition} outlines the key challenges and objectives guiding the development of our model. Sub-section \ref{sec:ResNetVLLM_model_formulation} details the architecture and mechanisms of ResNetVLLM, highlighting its unique features and improvements over existing methods. Sub-section \ref{sec:Zero-Shot-Video-Understanding-Strategy} explains our approach to achieving zero-shot video understanding.

\subsection{Problem Definition}
\label{sec:problem_definition}
\label{sec:training_process_definition}
\textbf{Input:} a video instance, denoted by \(v\).

\textbf{Output:} A natural language sentence \(s\), describing the content of video \(v\).

The task is to generate a sentence \(s\) that describe video \(v\).

\subsection{Training Process Definition}
\label{sec:training_process_definition}
\textbf{Input:} a set \(D\) of pairs (\(v, d\)). \(\mathcal{D} = \{v, d \mid v \in V, d \in D\}\), where:
\begin{itemize}
    \item \(V\) is a set of videos, where each video instance is denoted by \(v\).
    \item \(D\) is a set of video descriptions, where each description \(d\) corresponds to a video \(v\).
\end{itemize}

\textbf{Output:} A natural language sentence \(s\) for each video \(v\), describing its content.

The training task aims to train the ResNetVLLM model to generate sentences that describe videos from a training set \(D\), where \(\mathcal{D} = \{v, d \mid v \in V, d \in D\}\). Each video \(v\) in \(V\) is paired with its corresponding description \(d\) in \(D\). Our model learns to output a sentence \(s\) that describes the content of each video instance \(v\).

\subsection{ResNetVLLM Model Formulation}
\label{sec:ResNetVLLM_model_formulation}
ResNetVLLM consists of a visual encoder and a LLM transformer. We employ a 2D ResNet \cite{he2015deep} as the base network for visual feature extraction due to its efficiency in terms of memory and computation compared to 3D models. We begin by evenly selecting T frames from a video and using the ResNet visual encoder to extract features at the frame level. The extracted frame features are fed into the transformer network. Additionally, we integrate [CLS] tokens to collect global semantic features. To maintain computational efficiency, we use the flattened global features from the final average pooling layer of ResNet as the visual representations. This approach has proven effective in video understanding tasks \cite{wang2016temporal}. Lastly, the projected visual features are merged with the tokenized system command and fed into the LLM to produce the video description.

In the zero-shot setting, the ResNet feature encoder is initialized with random weights to prevent prior knowledge from influencing the model's performance. During training, both the ResNet feature encoder and the LLM transformer are optimized simultaneously to ensure cohesive learning and integration.

\subsection{Zero-Shot Video Understanding Strategy}
\label{sec:Zero-Shot-Video-Understanding-Strategy}
Our strategy for zero-shot video understanding leverages the model's ability to generalize from seen to unseen video content. We achieve this by training ResNetVLLM on a diverse set of video-text pairs, ensuring robust and transferable spatiotemporal and semantic representations. The fine-tuning process further enhances the model's adaptability to new and unseen video content, enabling it to generate accurate and contextually relevant textual descriptions even in zero-shot scenarios.

\begin{figure}[!t]
    \centering
    \includegraphics[width=0.30\textwidth,height=0.20\textheight]{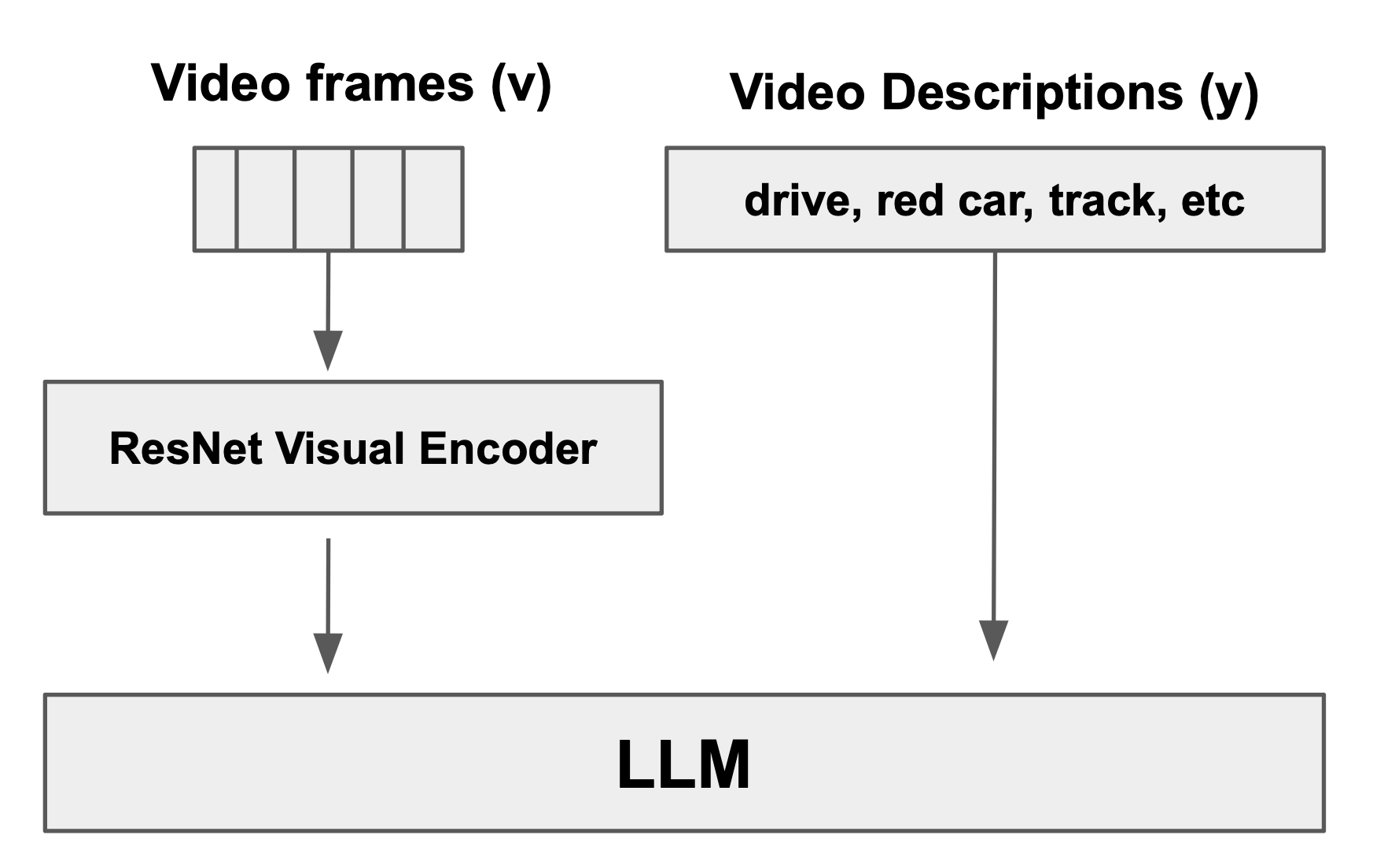}
    \caption{Training Phase of ResNetVLLM.}
    \label{fig:Training-Phase-figure}
\end{figure}

\section{Experimental evaluation}
\subsection{Experimental Setup}
\textbf{Datasets, training and evaluation protocol.} 
We trained our model using the Video-ChatGPT-100K instruction dataset \cite{maaz2023videochatgpt} and evaluated its performance on four video question-answering datasets: MSRVTT-QA \cite{xu2017video}, MSVD-QA \cite{xu2017video}, TGIF-QA FrameQA \cite{jang2017tgif}, and ActivityNet-QA \cite{yu2019activitynet}. The Video-ChatGPT-100K dataset provides diverse video-text pairs to facilitate robust training. 

To assess our model's performance in generating textual descriptions from video inputs, we use the VideoChatGPT benchmark \cite{maaz2023videochatgpt}, which includes two evaluation components:

\begin{enumerate}
  \item Video-based Generative Performance Benchmarking: This component measures the model's ability to generate coherent and contextually accurate text descriptions from video frames.
  \item Zero-Shot Question-Answer Evaluation: This evaluates the model's capability to answer questions about video content without prior training on specific question-answer pairs.
\end{enumerate}

\textbf{Implementation Details.} For each video, we select 100 frames at 6-frame intervals starting from a random point. Each frame is then randomly cropped to a 224 x 224 patch to standardize the input size. The ResNet visual encoder \cite{he2015deep} is adapted to process these video frames. In our experimental setup, we use one video clip for training and 25 clips for testing, ensuring a robust evaluation of the model's performance on unseen video content. The language model component of ResNetVLLM is based on LLaVA \cite{liu2023visual}, which integrates visual and textual information for improved understanding and generation.

Training is conducted in two stages: Warm-Up and Joint-Training. In the Warm-Up phase, the ResNet component is trained exclusively for 150 epochs using Stochastic Gradient Descent (SGD) with a learning rate of 0.01 and a weight decay of 0.0001. This phase focuses on optimizing the visual encoder to extract meaningful features from video frames. During this phase, Bayesian Optimization was employed to fine-tune critical hyperparameters, such as the learning rate and weight decay, by building a probabilistic model of the objective function and iteratively selecting the most promising hyperparameters. In the Joint-Training phase, the entire model, including both the ResNet encoder and the LLaVA transformer, is trained collectively for 50 epochs using the AdamW optimizer. This phase uses a learning rate of 0.00015 and a weight decay of 0.05 to fine-tune the model and improve overall performance. Bayesian Optimization was again utilized to optimize these parameters, ensuring that the model is both efficient and capable of handling complex video data. Training was performed on 2 Nvidia Tesla V100 GPUs, taking approximately 4 hours to complete.

\subsection{Evaluation Metrics}
\textbf{Video-based Text Generation Performance Benchmarking:} 
The first component of the VideoChatGPT benchmark \cite{maaz2023videochatgpt} is derived from the ActivityNet-200 dataset \cite{caba2015activitynet}, which comprises videos with detailed captions and human-annotated question-answer pairs. This evaluation framework, leveraging the GPT-3.5 model, evaluates the model's capabilities by assigning a relative score from 1 to 5 across five key areas:
\begin{enumerate}
    \item \textit{Correctness of Information (CI):} This criterion checks the accuracy of the generated text, ensuring it correctly reflects the video content without any misinterpretation or misinformation. CI is typically computed by comparing key factual elements in the generated text with ground truth information from the video. If the generated text contains factual inaccuracies (e.g., incorrect objects, actions, or properties), the score is reduced. Formula \ref{eq:correctness-of-information} shows how CI is computed.
    \begin{equation}
        \label{eq:correctness-of-information}
        \text{CI} = \frac{1}{N} \sum_{i=1}^{N} f_{\text{CI}}(A_i, \hat{A}_i)
    \end{equation}
    Where:
    \begin{itemize}
        \item \(N\) is the total number of videos in the dataset.
        \item \(f_{CI}\) is a function that evaluates the factual accuracy between the generated answer and the expected answer by dividing the number of correctly generated words by the total number of expected words.
        \item \(A_{i}\) is the correct answer for the \(i\)-th video.
        \item \(\hat{A}_i\) is the generated answer for the \(i\)-th video.
    \end{itemize}

    \item \textit{Detail Orientation (DO):} This aspect assesses the depth of the model's responses, ensuring they are comprehensive, covering all significant points from the video, and specific, including detailed rather than generic information. DO is assessed based on the inclusion of key details present in the video but often missed in generic descriptions. Object detection is used to identify the presence of detailed information (e.g., color, shape, texture, or other properties of objects). The equation for DO is show in \ref{eq:detail-orientation}
    \begin{equation}
        \label{eq:detail-orientation}
        \text{DO} = \frac{1}{N} \sum_{i=1}^{N} \left( w_c C_i + w_s S_i \right)
    \end{equation}
    Where:
    \begin{itemize}
        \item \(N\) is the total number of videos in the dataset.
        \item \(C_{i}\) is the completeness score for the \(i\)-th video.
        \item \(S_{i}\) is the specificity score for the \(i\)-th video.
        \item \(w_{c}\) and \(w_{s}\) are the weights for completeness and specificity, respectively.
    \end{itemize}

    \item \textit{Contextual Understanding (CU):} This factor evaluates the model's ability to understand and respond in alignment with the overall context of the video content. CU is evaluated by assessing the model's ability to generate text that reflects not only individual objects or actions but also their interactions and the broader scene in which they appear. This involves scene graph analysis to match the relationships between objects and events in the video with those described in the text.
    \begin{equation}
        \label{eq:contextual-understanding}
        \text{CU} = \frac{1}{N} \sum_{i=1}^{N} f_{\text{CU}}(Q, A, \hat{A})
    \end{equation}
    Where:
    \begin{itemize}
        \item \(N\) is the total number of videos in the dataset.
        \item \(f_{CU}\) is the function that evaluates the contextual understanding based on the comparison between the question, correct answer, and predicted answer.
    \end{itemize}
    {\scriptsize
    \[
    f_{\text{CU}}(Q, A, \hat{A}) = 
    \begin{cases} 
    5 & \text{if } \hat{A} \text{ fully aligns with } A \text{ in context} \\
    4 & \text{if } \hat{A} \text{ captures most themes of } A \\
    3 & \text{if } \hat{A} \text{ partially aligns with } A \text{ but misses key points} \\
    2 & \text{if } \hat{A} \text{ is somewhat out of context} \\
    1 & \text{if } \hat{A} \text{ has minimal alignment with } A \\
    0 & \text{if } \hat{A} \text{ is completely irrelevant to } A
    \end{cases}
    \]
    }

    \item \textit{Temporal Understanding (TU):} This measures the model's understanding of the sequence of events in the video when generating responses. TU is measured by checking if the sequence of events in the generated text matches the sequence in the video. Temporal order analysis can compare the order of events/actions between the generated text and the video.
    \begin{equation}
        \label{eq:temporal-understanding}
        \text{TU} = \frac{1}{N} \sum_{i=1}^{N} f_{\text{TU}}(Q, A, \hat{A})
    \end{equation}
    Where:
    \begin{itemize}
        \item \(N\) is the total number of videos in the dataset.
        \item \(f_{TU}\) is the function that evaluates predicted answers against correct answers, focusing on their temporal accuracy.
    \end{itemize}
    {\scriptsize
    \[
    f_{\text{TU}}(Q, A, \hat{A}) = 
    \begin{cases} 
    5 & \text{if } \hat{A} \text{ reflects the temporal sequence perfectly} \\
    4 & \text{if } \hat{A} \text{ captures most of the sequence accurately} \\
    3 & \text{if } \hat{A} \text{ shows some temporal alignment but misses} \\
      & \text{key events} \\
    2 & \text{if } \hat{A} \text{ is somewhat out of temporal order} \\
    1 & \text{if } \hat{A} \text{ has minimal alignment with the temporal order} \\
    0 & \text{if } \hat{A} \text{ is completely irrelevant to the temporal context}
    \end{cases}
    \]
    }

    \item \textit{Consistency (C):} This criterion examines the model's ability to maintain consistency across responses to similar questions or different sections of the video. Consistency is evaluated by identifying contradictions, incoherence, or repetition in the generated text. It is calculated based on logical coherence models or by checking if facts presented early in the text are consistent with those presented later.
    \begin{equation}
        \label{eq:consistency}
        \text{C} = \frac{1}{N} \sum_{i=1}^{N} f_{\text{C}}(Q_1, Q_2, A, \hat{A}_1, \hat{A}_2)
    \end{equation}
    Where:
    \begin{itemize}
        \item \(N\) is the total number of videos in the dataset.
        \item \(f_{C}\) is the function that evaluates the consistency of predicted answers for two similar questions, focusing on their alignment with each other and the correct answer.
    \end{itemize}
    {\scriptsize
    \[
    f_{\text{C}}(Q_1, Q_2, A, \hat{A}_1, \hat{A}_2) = 
    \begin{cases} 
    5 & \text{if } \hat{A}_1 \text{ and } \hat{A}_2 \text{ are completely consistent} \\
      & \text{with each other and with } A \\
    4 & \text{if } \hat{A}_1 \text{ and } \hat{A}_2 \text{ are mostly consistent with} \\
      & \text{minor differences} \\
    3 & \text{if } \hat{A}_1 \text{ and } \hat{A}_2 \text{ show some inconsistencies} \\
      & \text{but are generally aligned with } A \\
    2 & \text{if } \hat{A}_1 \text{ and } \hat{A}_2 \text{ are inconsistent with} \\
      & \text{each other} \\
    1 & \text{if } \hat{A}_1 \text{ or } \hat{A}_2 \text{ has minimal alignment} \\
      & \text{with } A \\
    0 & \text{if } \hat{A}_1 \text{ and } \hat{A}_2 \text{ contradict each other or} \\
      & \text{are irrelevant to } A
    \end{cases}
    \]
    }
    
    \item \textit{Mean:} This is the overall performance score. It is calculated by averaging the scores from the five above evaluation metrics. \begin{equation}\text{Mean} = \frac{\text{CI + DO + CU + TU + C}}{\text{5}}\end{equation}
\end{enumerate}

\textbf{Zero-Shot Question-Answer Evaluation:}
The second component of the VideoChatGPT benchmark \cite{maaz2023videochatgpt} involves a thorough quantitative assessment using well-established open-ended question-answer datasets: MSRVTT-QA \cite{xu2017video}, MSVD-QA \cite{xu2017video}, TGIF-QA FrameQA \cite{jang2017tgif}, and ActivityNet-QA \cite{yu2019activitynet}. These assessments were conducted in a zero-shot scenario, employing GPT-guided evaluation to analyze the model’s accuracy. In the context of LLM models, accuracy refers to the model's ability to generate outputs that are correct, relevant, and aligned with the task or prompt given. It can be measured in different ways depending on the application. Accuracy in the contexts of video understanding refers to how well the model correctly identifies the scene context or environment depicted in the video. Formula \ref{eq:accuracy-in-zsqa} shows how \text{accuracy in ZSQA} is calculated.

\begin{equation}
    \label{eq:accuracy-in-zsqa}
    \text{Accuracy in ZSQA} = \frac{1}{N} \sum_{i=1}^{N} C_i
\end{equation}
Where:
\begin{itemize}
    \item \(N\) is the total number of videos in the dataset.
    \item \(C_{i}\) represents the number of correct answers for the \(i\)-th video.
\end{itemize}

\begin{table}[!h]
    \renewcommand{\arraystretch}{1.25} % Increase the space between rows by 25%
    \centering
    \caption{Comparison with recent state-of-the-art methods on the \textbf{Video-based Text Generation Performance Benchmarking} component of the video understanding benchmark \cite{maaz2023videochatgpt}. The benchmark pipeline assesses various model capabilities and assigns a rating on a scale of 1-5 to the generated predictions across the following five aspects.}
    \label{tab:video-based-text-generation-performance}
    \begin{tabular}{lllllll}
        \hline
        \textbf{Method} & \textbf{CI} & \textbf{DO} & \textbf{CU} & \textbf{TU} & \textbf{C} & \textbf{Mean} \\
        \hline
        Video Chat \cite{li2023videochat} & 2.25 & 2.50 & 2.54 & 1.98 & 1.84 & 2.22 \\
        LLaMA Adapter v2 \cite{gao2023llamaadapter} & 2.03 & 2.32 & 2.30 & 1.98 & 2.15 & 2.16 \\
        Video LLaMA \cite{zhang2023videollama} & 1.96 & 2.18 & 2.16 & 1.82 & 1.79 & 1.98 \\
        Video-ChatGPT \cite{maaz2023videochatgpt} & 2.50 & 2.57 & 2.69 & 2.16 & 2.20 & 2.42 \\
        Valley \cite{luo2023valley} & 2.43 & 2.13 & 2.86 & 2.04 & 2.45 & 2.38 \\
        BT-Adapter \cite{liu2023one} & 2.16 & 2.46 & 2.89 & 2.13 & 2.20 & 2.37 \\
        \hline
        \textbf{ResNetVLLM (ours)} & \textbf{3.64} & \textbf{3.19} & \textbf{3.92} & \textbf{3.16} & \textbf{3.84} & \textbf{3.55} \\
        \hline
    \end{tabular}
\end{table}

\subsection{Main Results}
\begin{table*}
    \centering
    \setlength{\tabcolsep}{6pt}
    \renewcommand{\arraystretch}{1.25} % Increase the space between rows by 25%
    \caption{Comparison with recent state-of-the-art methods on the \textbf{The Zero-Shot Question-Answer Evaluation} component of the video understanding benchmark \cite{maaz2023videochatgpt} utilizes GPT to assess the accuracy of the model's generated responses. Accuracy in the contexts of video understanding refers to how well the model correctly identifies the scene context or environment depicted in the video.}
    \label{tab:zero-shot-question-answer-evaluation}
    \resizebox{\textwidth}{!}{
    \begin{tabular}{l c c c c}
        \hline
         \textbf{Model} & \textbf{MSVD-QA} & \textbf{MSRVTT-QA} & \textbf{TGIF-QA} & \textbf{Activity Net-QA} \\
        \hline
        FrozenBiLM \cite{yang2022zero} & 32.2 & 16.8 & 41.0 & 24.7 \\
        Video Chat \cite{li2023videochat} & 56.3 & 45.0 & 34.4 & 26.5 \\
        LLaMA Adapter v2 \cite{gao2023llamaadapter} & 54.9 & 43.8 & - & 34.2 \\
        Video LLaMA \cite{zhang2023videollama} & 51.6 & 29.6 & - & 12.4 \\
        Video-ChatGPT \cite{maaz2023videochatgpt} & 64.9 & 49.3 & 51.4 & 35.2 \\
        Valley \cite{luo2023valley} & 60.5 & 51.1 & - & 45.1 \\
        BT-Adapter \cite{liu2023one} & 67.5 & 57.0 & - & 45.7 \\
        \hline
        \textbf{ResNetVLLM (ours)} & \textbf{78.3} & \textbf{63.5} & \textbf{59.9} & \textbf{54.8} \\
        \hline
    \end{tabular}}
\end{table*}

\textbf{Video-based Text Generation Performance Benchmarking:} We report the evaluation results of our proposed model, ResNetVLLM as depicted in Table\ref{tab:video-based-text-generation-performance}. The results demonstrate its competent performance across all key aspects compared to recent state-of-the-art methods. We attribute the improvement in performance to the minimized information loss in our integrated cross-modal framework, which effectively leverages visual distinctions for efficient knowledge transfer. This enhancement equips ResNetVLLM with a strong capability to produce contextually appropriate, detailed, and temporally precise text from video input.

\textbf{Zero-Shot Question-Answer Evaluation:} To evaluate our model, we compared its accuracy against other notable models, such as FrozenBiLM \cite{yang2022zero}, and the generative video models Video Chat \cite{li2023videochat}, LLaMA Adapter \cite{gao2023llamaadapter}, Video LLaMA \cite{zhang2023videollama}, and Video-ChatGPT \cite{maaz2023videochatgpt}. Despite the robust capabilities of these models, ResNetVLLM consistently outperformed them, achieving state-of-the-art performance across all datasets. These findings underscore ResNetVLLM's ability to understand video content and produce precise, contextually nuanced responses to queries. The results are displayed in Table \ref{tab:zero-shot-question-answer-evaluation}

\section{Conclusion}
ResNetVLLM represents an advancement in the field of video understanding, leveraging a unique combination of a vanilla ResNet visual encoder and an LLM to achieve exceptional performance in zero-shot scenarios. Unlike previous methods that rely on pre-trained video understanding models or manual labeling, ResNetVLLM's approach to visual feature extraction and semantic integration within a single framework provides a more robust and adaptable solution for generating textual descriptions from video content. The comprehensive evaluation of ResNetVLLM across multiple benchmarks confirms its superiority over existing state-of-the-art models, highlighting its capability to understand and interpret complex video data effectively. This paper's contributions underscore the potential of integrating visual and semantic learning in a unified model, paving the way for further advancements in video-based language tasks and zero-shot learning.

\bibliographystyle{IEEEtran}
\bibliography{IEEEabrv,paper}

\appendices
\onecolumn % Switch to single column layout for appendices
\section{Sample Output of ResNetVLLM}
\label{app:A-Appendix}
\begin{figure}[h]
    \centering
    \includegraphics[width=\textwidth,height=0.30\textheight]{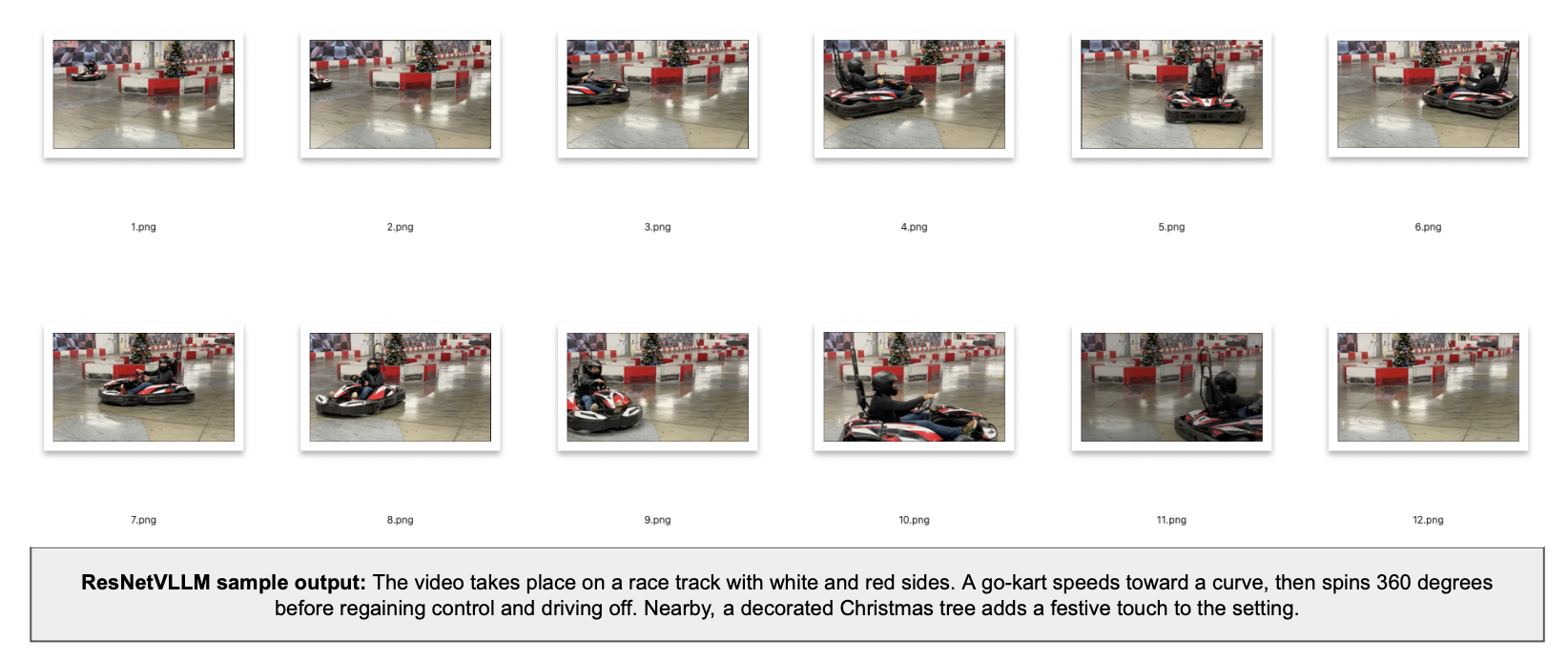}
    \caption{Sample Output of ResNetVLLM.}
    \label{fig:ResNetVLLM-sample-output-2}
\end{figure}

\end{document}